%% file: iclr2026_workshop.tex
\documentclass{article} 
\usepackage{iclr2026_workshop,times}

\input{math_commands.tex}

\usepackage{hyperref}
\usepackage{url}
\usepackage{booktabs}
\usepackage{capt-of}
\usepackage{graphicx}
\usepackage{multirow}

\title{CA-BED: Conversation-Aware Bayesian Experimental Design}

\author{Daniel Arnould \\
\texttt{daniel.s.arnould@gmail.com} \\
\And
Rashad Aziz \\
\texttt{rashadaziz.p@gmail.com} \\
\And{Zixuan Kang}
\And{Tanav Changal}
\And{Kevin Zhu}
\And{Sunishchal Dev}
\And{Gabriel Grand}
\And{Shreyas Sunil Kulkarni}
}

\iclrfinaltrue

\begin{document}

\maketitle

\begin{abstract}

Large Language Models (LLMs) excel at static reasoning tasks, yet their
performance often degrades in interactive scenarios where information must be
actively acquired through questioning. A key challenge lies in selecting
questions that reduce uncertainty while incorporating responses that may be
ambiguous or only partially informative. To address this, we propose
Conversation-Aware Bayesian Experimental Design
(CA-BED)\footnote{Code is available at \url{https://github.com/DanielArnould/ca-bed}.},
an inference-time probabilistic dialog planning framework that integrates
Bayesian Experimental Design with LLM-based likelihood estimation to optimize
question selection over multiple conversational turns. CA-BED maintains a
belief distribution over hypotheses, anticipates possible answers, and
propagates expected information gain through a simulated conversation tree.
Across two structured entity-deduction benchmarks, CA-BED yields an average
21.8\% improvement in success rates over direct prompting, with comparable
gains relative to alternative information-seeking methods. It achieves these
gains with an average increase of only 1.8 conversational turns compared to
direct prompting.
These results suggest that probabilistic conversation planning is a promising
direction for interactive reasoning in structured information-seeking settings.
\end{abstract}

\section{Introduction}

Large Language Models (LLMs) excel at static reasoning tasks such as code
generation and logic puzzles \citep{one, two, wang2025surveylargelanguagemodels,
jiang2024surveylargelanguagemodels}, yet their performance declines in
interactive settings requiring active information gathering. Persistent
weaknesses appear in entity-deduction \citep{zhang2023entity,
zhou2025passiveactivereasoninglarge, bertolazzi-etal-2023-chatgpts}, tool use
\citep{patil2025bfcl}, and clarification tasks
\citep{gan2024clarqllmbenchmarkmodelsclarifying}. In domains like tutoring,
diagnosis, and troubleshooting, effective questioning is essential; models must
decide not only how to reason, but when and what to ask. Because exhaustively
defining all possible queries is infeasible, adaptive information acquisition
remains a key challenge for reliable LLM deployment.

Recent approaches address this through (1) fine-tuning models to generate
questions \citep{collabllm2025, zhang2025modelingfutureconversationturns,
andukuri2024stargateteachinglanguagemodels, li2025alfaaligningllmsask}, which
often struggle to generalize, and (2) training-free planning methods where
models simulate dialog trajectories before acting
\citep{hu2024uncertaintythoughtsuncertaintyawareplanning,
yu2023promptbasedmontecarlotreesearch, li2024mediqquestionaskingllmsbenchmark}.
The Uncertainty of Thoughts (UoT) framework
\citep{hu2024uncertaintythoughtsuncertaintyawareplanning} treats conversation as
a decision process over a contextual tree, selecting questions that partition
the hypothesis space efficiently.

However, UoT and similar methods
\citep{chan2025conformalinformationpursuitinteractively,
chopra2025feedbackawaremontecarlotree} falter with partial or ambiguous
evidence. Real-world queries like “Do you have a fever?” should adjust, rather
than eliminate, hypotheses. Allowing soft partitions mitigates this but often
extends dialog unnecessarily. A principled approach is needed to represent
graded evidence and propagate uncertainty through the conversation.

We propose \textbf{Conversation-Aware Bayesian Experimental Design (CA-BED)},
which integrates Bayesian Experimental Design (BED) with conversational planning
to enhance efficiency and robustness in LLM-based information seeking. BED
optimizes data acquisition by maximizing expected information gain \citep{982896,
Berry2006, choudhury2025bedllmintelligentinformationgathering}; CA-BED extends
this to dialog, treating each question as a Bayesian experiment that updates
beliefs over hypotheses.

\begin{center}
\includegraphics[width=\textwidth]{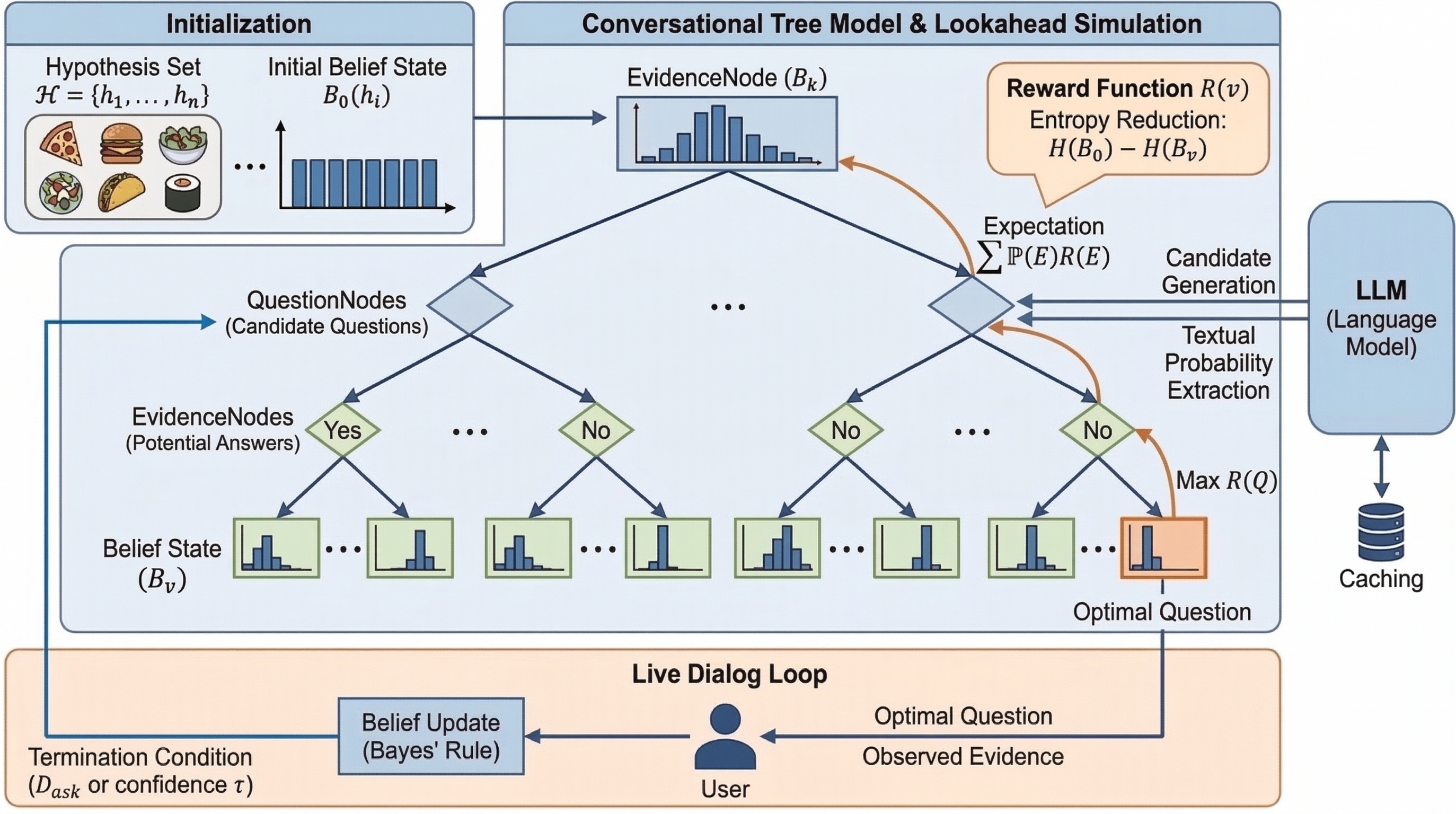}
\captionof{figure}{Overview of CA-BED. The agent maintains a belief distribution over candidate hypotheses, generates candidate questions, expands an alternating QuestionNode/EvidenceNode conversation tree over possible answers, and estimates answer likelihoods with an LLM to score branches by expected information gain. After observing the real user response, the posterior belief is updated and the process repeats until the confidence threshold is met or the conversation budget is exhausted.}
\label{fig:main_overview}
\end{center}

\textbf{Contributions:}
\begin{itemize}
\item \textbf{Uncertainty-aware conversation planning:} A hybrid framework
combining UoT with probabilistic belief modeling and LLM-based likelihood
estimation for dynamic uncertainty reasoning.
\item \textbf{Multi-answer planning:} Anticipating diverse responses improves
information efficiency and robustness, capturing the ambiguity of real dialog.
\item \textbf{Empirical validation:} CA-BED outperforms prior baselines on
\textit{Detective Cases} and \textit{20 Questions}, improving both accuracy and
conversational efficiency.
\end{itemize}

\section{Related Work}

Research on LLM reasoning has focused primarily on static tasks like math and
logic \citep{wang2025surveylargelanguagemodels,
lin2025zebralogicscalinglimitsllms, xu2025largereasoningmodelssurvey}, while
interactive reasoning, where models actively seek information, remains
underexplored. Prior work distinguishes clarification tasks, which resolve
ambiguous prompts \citep{tanjim2025disambiguationconversationalquestionanswering,
Aliannejadi_2019, deng2023promptingevaluatinglargelanguage}, from
information-seeking tasks, which study how LLMs query or use tools to obtain
missing facts \citep{handa2024bayesianpreferenceelicitationlanguage,
patil2025bfcl}.

Information-seeking benchmarks often use entity-deduction games like Twenty
Questions \citep{bertolazzi-etal-2023-chatgpts, zhang2023entity}, Guess the
Celebrity \citep{choudhury2025bedllmintelligentinformationgathering}, and medical
diagnosis \citep{li2024mediqquestionaskingllmsbenchmark}. Larger setups such as
AR-bench \citep{zhou2025passiveactivereasoninglarge} extend this with multiple
speaking agents and show that LLMs still lag behind humans.

Some works fine-tune models to ask better questions \citep{collabllm2025,
zhang2025modelingfutureconversationturns}, while others use online reasoning
guided by heuristics like information gain or entropy
\citep{chan2025conformalinformationpursuitinteractively,
choudhury2025bedllmintelligentinformationgathering,
hu2024uncertaintythoughtsuncertaintyawareplanning,
wang2025adaptiveelicitationlatentinformation,
cooper2025curiouslanguagemodelstrategic}. One influential approach in this domain 
is the Uncertainty of Thoughts (UoT) framework \citep{hu2024uncertaintythoughtsuncertaintyawareplanning}, 
which structures the information-seeking process as a simulated dialog tree. UoT 
actively searches for a sequence of binary ``Yes/No'' questions designed to rapidly 
partition the space of possible entities. While effective in constrained entity-deduction 
scenarios with small, well-defined hypothesis sets, this reliance on hard, binary 
partitioning causes the method to struggle when navigating large search spaces or 
processing ambiguous, real-world answers \citep{zhou2025passiveactivereasoninglarge}.

Many of these approaches can be seen as special cases of Bayesian Experimental
Design (BED), which selects actions that maximize expected information gain
\citep{982896, Berry2006, rainforth2023modernbayesianexperimentaldesign}. Recently, 
\citet{choudhury2025bedllmintelligentinformationgathering} formalized this connection 
through BED-LLM, establishing that each generated question acts as an experiment 
designed to isolate an answer from a broader hypothesis space, with likelihood updates 
derived either implicitly or explicitly. Our work, CA-BED, builds directly upon this 
formalization with a practical implementation that uses an LLM to extract likelihood updates. Furthermore, rather than just optimizing the next question, we introduce a look-ahead  to propagate expected information gain across longer, 
multi-turn conversation trajectories. Lastly, we assess the impact of anticipating multiple responses instead of the typical binary answers.

\section{Methodology}

In this section, we present the Conversation-Aware Bayesian Experimental Design (CA-BED) algorithm. The goal of CA-BED is to optimize the problem of entity-deduction conversations, where an agent must converse with a user to extract an answer from some hypothesis set (e.g., a doctor trying to identify a patient's symptoms). This covers a wide class of important scenarios where information acquisition is required, such as interactive troubleshooting, customer support triage, and adaptive educational tutoring.

\subsection{Probabilistic Model}

The algorithm operates by framing the conversational information-seeking process as a sequential decision problem under uncertainty. We define a discrete, finite set of mutually exclusive hypotheses as $\mathcal{H} = \{h_1, \dots, h_n\}$. At any step $k$ in the dialog, the agent maintains a belief state, defined as the posterior probability distribution over the hypothesis set given the sequence of observed evidence (answers) $E_{1:k}$:
\[ B_k(h_i) = P(h_i|E_{1:k}) \]

The conversation begins with an initial belief state $B_0(h_i) = P(h_i)$. This prior distribution can be strictly uniform, assuming that all hypotheses are equally likely, or it can be initialized with an informative prior if domain-specific context is available.

Observe that this formalization requires an \textit{a priori} defined hypothesis space, which inherently restricts fully open-domain conversations. However, for entity-deduction tasks, a taxonomy of the hypothesis set is often available. Furthermore, this restriction does not necessarily mean the hypothesis set is permanently fixed; it can be dynamically generated or bounded by an LLM at the beginning of the interaction and refined as new, unexpected information emerges.

\subsubsection{Conversational Tree Model}

Since the problem is formulated as a decision process, we can now visualize and compute the forward trajectory of a dialog as an alternating decision tree. This tree is composed of two distinct node types:

\begin{itemize}
    \item \textbf{QuestionNodes:} These represent the active interventions by the model to refine the hypothesis space.
    \item \textbf{EvidenceNodes:} These represent the resulting observations. Each possible answer from the user branches into a unique EvidenceNode, containing the updated belief state $B_k$.
\end{itemize}

In this tree structure, nodes alternate strictly: every QuestionNode branches into multiple EvidenceNodes (representing possible answers), and every EvidenceNode can branch into subsequent QuestionNodes (representing follow-up queries).

\subsection{Optimal Question Selection}
To navigate this tree, we define a reward function, $R(v)$ for any given node $v$. This reward function is defined recursively based on the node type:
\[
R(v) = \begin{cases}
  \sum_{E \in \text{Children}(v)} \mathbb{P}(E) R(E) & \text{if } v \text{ is a QuestionNode}, \\
  \max_{Q \in \text{Children}(v)} R(Q) & \text{if } v \text{ is an internal EvidenceNode}, \\
  H(B_0) - H(B_v) & \text{if } v \text{ is a terminal EvidenceNode}
\end{cases}
\]
For a QuestionNode, the reward is the expected value of its children, weighted by the marginal probability $\mathbb{P}(E)$ of actually receiving that specific answer. This penalizes highly specific questions that yield high information gain if successful, but have a near-zero probability of occurring.

For an internal EvidenceNode, the reward is the maximum reward of its available follow-up questions. This ensures that we evaluate the actual ceiling of the optimal trajectory, as we would never pursue a worse branch. Considering neighboring branches, such as by taking the mean of their rewards, could be less effective as it would artificially penalize a highly optimal conversational path simply because the agent also happened to generate weak, uninformative sibling branches.

For a terminal EvidenceNode, the reward is defined as the reduction in Shannon entropy between the initial belief state and the belief state at node $v$. The Shannon entropy of a belief distribution $B$ is given by
\[
H(B) = -\sum_{h \in \mathcal{H}} B(h)\log_2 B(h),
\]
which quantifies the uncertainty over the hypothesis space. As is standard, zero probabilities are ignored from this sum \citep{rainforth2023modernbayesianexperimentaldesign}. Higher entropy corresponds to greater uncertainty, while lower entropy indicates a more concentrated belief distribution.

By taking the difference $H(B_0) - H(B_v)$, the reward measures the total reduction in uncertainty achieved along a trajectory, which corresponds to the information gain. This formulation is consistent with the standard objective in Bayesian Experimental Design \citep{rainforth2023modernbayesianexperimentaldesign}, where the goal is to select actions that maximize the expected reduction in posterior uncertainty. Entropy provides a principled and model-agnostic measure of uncertainty, making it well-suited for comparing candidate questions under different possible outcomes.

In a live dialog, the agent selects the question that maximizes this reward and queries the user accordingly. The interaction proceeds iteratively, updating the belief state after each observation, until a termination condition is met. Specifically, the dialog terminates when either the maximum conversation depth $D_{\text{ask}}$ is reached or the posterior probability of any hypothesis $h_i$ exceeds a predefined confidence threshold $\tau$.

\subsection{Tree Construction}
During a live dialog, we build this tree and select the optimal path dynamically with the actual user.

\textbf{Candidate Generation:} We prompt an LLM to generate $M_q$ candidate questions conditioned on the current dialog history and remaining viable hypotheses.

\textbf{Likelihood Estimation:} For each question, we estimate the likelihood of observing evidence $E$ given a hypothesis, $\mathbb{P}(E|h_i)$. We acquire these likelihoods by explicitly prompting the LLM to output numerical probabilities directly as text, which we then extract. While extracting token log-probabilities might seem more direct, modern instruction-tuned LLMs either do not expose these or they are heavily collapsed, requiring temperature calibration to get useful results. Additionally, textually extracting probabilities is often used in literature with better performance than log-probabilities \citep{nan2026interpretableprobabilityestimationllms}. To account for epistemic uncertainty (e.g., a user giving an ambiguous or slightly inaccurate answer), we also apply a confidence-weighted smoothing to these likelihoods, pulling them toward a uniform baseline $\mathbb{P}'(E_k|h_i) = \epsilon \mathbb{P}(E_k|h_i) + (1-\epsilon)$.

\textbf{Belief Update:} We compute the transition to new EvidenceNodes via Bayes' rule, assuming conditional independence of the evidence given $h_i$. This Naive Bayes assumption is standard in dynamic probabilistic modeling to maintain computational tractability \citep{naivebayes}; it posits that once the ground-truth hypothesis is known, individual pieces of observed evidence do not strictly depend on one another, allowing for simple sequential belief updates.

\begin{center}
\centering
\includegraphics[width=0.84\textwidth]{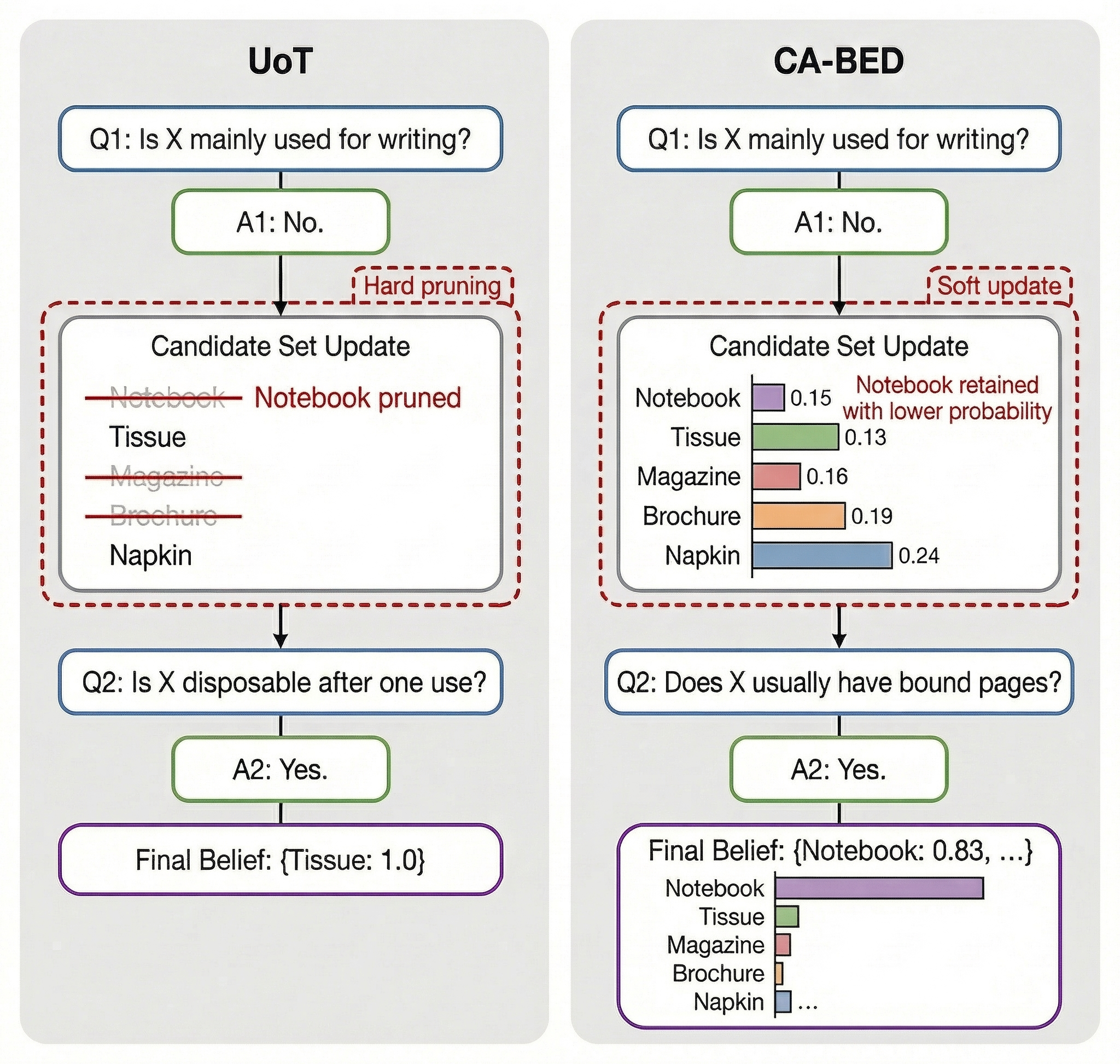}
\captionof{figure}{Representative example of belief updating on an ambiguous exchange from \textit{20 Questions}. Under hard pruning, a plausible hypothesis can be removed after a single imperfect answer. In contrast, CA-BED down-weights that hypothesis while preserving uncertainty, allowing a later question to recover the correct candidate.}
\label{fig:soft_update_example}
\end{center}

\textbf{Lookahead:} For each answer, we recursively simulate the conversation trajectory $D_{\text{sim}}$ steps further. While expanding this lookahead tree introduces exponential computational complexity---scaling at $\mathcal{O}((M_q \cdot K)^{D_{\text{sim}}})$ where $K$ is the number of possible answers---we can mitigate this significantly. Because the likelihood $\mathbb{P}(E_k|h_i)$ is strictly dependent on the isolated text of the question and the definition of the hypothesis, it is conditionally independent of the dialog history leading up to that node. Consequently, we can aggressively cache these LLM likelihood queries. As subtrees frequently explore similar semantic concepts or reuse optimal questions, this caching mechanism reduces the empirical API footprint to a computationally tractable level.

\section{Experimental Setup}
\subsection{Research Questions}
We aim to evaluate the effectiveness of CA-BED for improving information acquisition in conversational reasoning. 
Specifically, our experiments address the following questions:

\begin{itemize}
    \item \textbf{RQ1:} Does CA-BED increase information acquisition over zero-shot prompting?
    \item \textbf{RQ2:} Do explicit likelihoods offer better information acquisition than UoT?
    \item \textbf{RQ3:} Can answer planning improve information acquisition?
    \item \textbf{RQ4:} Does CA-BED increase information acquisition even in unsuccessful tasks?
\end{itemize}

\subsection{Tasks \& Benchmarks}
We evaluate the above questions on two practical entity-deduction scenarios:

\begin{itemize}
    \item \textbf{20 Questions:} A classic guessing game in which one player asks up to twenty questions to identify a hidden entity selected by another player. We use the COMMON dataset from \citet{hu2024uncertaintythoughtsuncertaintyawareplanning}, consisting of 111 animals, objects, places, and foods.
    \item \textbf{Detective Cases:} A murder mystery game introduced by \citet{zhou2025passiveactivereasoninglarge} of 100 different scenarios. 
An agent is provided with a scenario and five suspects, and can interrogate each suspect to deduce the murderer.
\end{itemize}

For both datasets, we test every sample asynchronously over 3 random seeds. This means that results found in one sample can be reused across samples, which is particularly relevant for \textit{20 Questions}.

\subsection{Compared Methods}
To analyze CA-BED's contributions, we compare four configurations. Across all configurations, we use \textbf{DeepSeek-V3.2 Chat} as both the Questioner and the Answerer, and we set the sampling temperature to $1.0$ for all generation tasks. A supplementary ablation over alternative Questioner models is reported in Appendix~\ref{sec:appendix_questioners}.

\begin{itemize}
    \item \textbf{Direct Prompting:} 
    The Questioner LLM receives the conversation history and generates either another question or a final prediction.

    \item \textbf{Uncertainty of Thoughts (UoT):} 
    A special case of CA-BED reflecting prior baselines \citep{hu2024uncertaintythoughtsuncertaintyawareplanning} where the estimator confidence is $\epsilon = 1.0$ and the confidence threshold is $\tau = 1.0$. 
    Each answer has a binary likelihood ($1$ or $0$). 
    We set the branching width to $M_q = 3$ and the lookahead depth to $D_{\text{sim}} = 2$.

    \item \textbf{CA-BED:} 
    Our proposed method, using a branching width of $M_q = 3$ and a lookahead depth of $D_{\text{sim}} = 2$. We set the estimator confidence to $\epsilon = 0.7$ and the confidence threshold to $\tau = 0.8$. 
    The Questioner is restricted to binary (``yes''/``no'') questions

    \item \textbf{CA-BED + Answer Planning:} 
    An extension of CA-BED using the exact same hyperparameters ($M_q = 3$, $D_{\text{sim}} = 2$, $\epsilon = 0.7$, $\tau = 0.8$). However, instead of being restricted to binary options, answer planning works by framing questions that are not binary, allowing the Questioner to generate and evaluate multiple candidate answers dynamically.
\end{itemize}

\section{Results \& Discussion}
\subsection{RQ1: Effectiveness of CA-BED}

As shown in Table~\ref{tab:main_results}, CA-BED consistently outperforms direct prompting across both benchmarks. In \textit{Detective Cases}, it achieves a $46.0\%$ success rate compared to direct prompting's $27.6\%$, exchanging a modest increase in conversation length for a more robust deduction process. This advantage is even more pronounced in \textit{20 Questions}, where CA-BED yields a $25.2\%$ absolute improvement in accuracy over the zero-shot baseline.

\begin{center}
\centering
\captionof{table}{Performance comparison across Detective Cases and 20 Questions. Success Rate (SR) includes Wilson's 95\% Confidence Intervals [in brackets]. Mean Conversation Length (MC) and Mean Successful Conversation Length (MSC) are reported with the Standard Error of the Mean ($\pm$ SEM).}
\label{tab:main_results}
\resizebox{\textwidth}{!}{%
\begin{tabular}{lcccccc}
\toprule
& \multicolumn{3}{c}{Detective Cases} & \multicolumn{3}{c}{20 Questions} \\
\cmidrule(lr){2-4} \cmidrule(lr){5-7}
Method & SR$\uparrow$ & MC$\downarrow$ & MSC$\downarrow$ & SR$\uparrow$ & MC$\downarrow$ & MSC$\downarrow$ \\
\midrule
Direct Prompting         & 27.6 [19.7, 37.1]                   & 7.4 $\pm$ 0.18          & 6.0 $\pm$ 0.34          & 57.7 [48.4, 66.4]                   & 9.7 $\pm$ 0.38          & 7.7 $\pm$ 0.27 \\
UoT                      & 29.0 [21.0, 38.5]                   & \textbf{2.7 $\pm$ 0.16} & \textbf{2.4 $\pm$ 0.28} & 60.4 [51.1, 69.0]                   & \textbf{7.2 $\pm$ 0.13} & \textbf{7.1 $\pm$ 0.13} \\
CA-BED                   & \textbf{46.0} [\textbf{36.6, 55.7}]                   & 7.9 $\pm$ 0.24          & 7.6 $\pm$ 0.34          & 82.9 [74.8, 88.8]                   & 12.8 $\pm$ 0.40         & 11.7 $\pm$ 0.34 \\
CA-BED + Answer-Planning & 38.0 [29.1, 47.8] & 8.9 $\pm$ 0.20          & 9.3 $\pm$ 0.25          & \textbf{89.2} [\textbf{82.0, 93.7}] & 8.8 $\pm$ 0.33          & 8.1 $\pm$ 0.22 \\
\bottomrule
\end{tabular}%
}
\end{center}

Compared to UoT, CA-BED produces significant absolute gains in success rate ($+17.0\%$ in \textit{Detective Cases} and $+22.5\%$ in \textit{20 Questions}) but requires longer interactions. This reflects UoT's strategy of aggressively halving the hypothesis space with binary hard-eliminations, which converges rapidly (e.g., $2.7$ turns in \textit{Detective Cases}) but frequently falters under ambiguous evidence. In contrast, CA-BED's soft probabilistic belief updates grant it resilience against uncertainty at the cost of a longer interaction span.

\subsection{RQ2: Do Explicit Likelihoods Improve Information Acquisition Over UoT?}

To examine whether the performance gap between UoT and CA-BED can be mitigated by incorporating epistemic uncertainty, UoT was extended with a probability-based belief state. The original binary partitioning of the hypothesis space was retained, but likelihood estimates were smoothed using an estimator confidence parameter. Specifically, rather than assigning deterministic likelihoods (e.g., setting $P(\text{Yes} \mid h) = 1.0$), values are smoothed toward a uniform baseline using an estimator confidence of $\epsilon = 0.7$. This modified variant was evaluated under different termination thresholds, $\tau \in \{0.8, 1.0\}$, to assess the interaction between likelihood smoothing and stopping criteria.

\begin{center}
\centering
\captionof{table}{Comparison of CA-BED and UoT augmented with epistemic uncertainty. Values in parentheses indicate $(\epsilon, \tau)$, representing the estimator confidence and confidence threshold parameters, respectively.}
\label{tab:soft_uot}
\resizebox{\textwidth}{!}{%
\begin{tabular}{lcccccc}
\toprule
& \multicolumn{3}{c}{Detective Cases} & \multicolumn{3}{c}{20 Questions} \\
\cmidrule(lr){2-4} \cmidrule(lr){5-7}
Method & SR$\uparrow$ & MC$\downarrow$ & MSC$\downarrow$ & SR$\uparrow$ & MC$\downarrow$ & MSC$\downarrow$ \\
\midrule
UoT (1.0, 1.0)    & 29.0 [21.0, 38.5]                   & \textbf{2.7 $\pm$ 0.16} & \textbf{2.4 $\pm$ 0.28} & 60.4 [51.1, 69.0]                   & \textbf{7.2 $\pm$ 0.13} & \textbf{7.1 $\pm$ 0.13} \\
UoT (0.7, 1.0)    & 41.0 [31.9, 50.8]                   & 10.0 $\pm$ 0.0          & 10.0 $\pm$ 0.0          & 89.2 [82.0, 93.7]                   & 19.8 $\pm$ 0.17         & 20.0 $\pm$ 0.0 \\
UoT (1.0, 0.8)    & 29.0 [21.0, 38.5]                   & \textbf{2.7 $\pm$ 0.16} & \textbf{2.4 $\pm$ 0.28} & 60.4 [51.1, 69.0]                   & \textbf{7.2 $\pm$ 0.13} & \textbf{7.1 $\pm$ 0.13} \\
UoT (0.7, 0.8)    & 33.0 [24.6, 42.7]                   & 5.3 $\pm$ 0.26          & 5.8 $\pm$ 0.49          & \textbf{90.1} [\textbf{83.1, 94.4}] & 12.8 $\pm$ 0.37         & 12.3 $\pm$ 0.32 \\
CA-BED (0.7, 0.8) & \textbf{46.0} [\textbf{36.6, 55.7}] & 7.9 $\pm$ 0.24          & 7.6 $\pm$ 0.34          & 82.9 [74.8, 88.8]                   & 12.8 $\pm$ 0.40         & 11.7 $\pm$ 0.34 \\
\bottomrule
\end{tabular}%
}
\end{center}

The results in Table~\ref{tab:soft_uot} highlight a strict operational dependency between these parameters. When smoothing is applied ($\epsilon = 0.7$) but the termination threshold remains absolute ($\tau = 1.0$), the dampened likelihoods prevent the belief state from ever satisfying the stopping criterion. Consequently, the agent exhausts the conversation budget, reflected in mean conversation lengths approaching the maximum allowed depth (10.0 turns for \textit{Detective Cases} and 19.8 turns for \textit{20 Questions}).

Lowering the threshold to $\tau = 0.8$ aligns it with the effective likelihood ceiling, reducing mean conversation lengths to 5.3 and 12.8 turns, respectively. Under this configuration, the smoothed UoT variant shows substantial improvements over its deterministic counterpart. On \textit{20 Questions}, Top-1 accuracy increases from 60.4\% to 90.1\%, matching CA-BED's performance. This suggests that in simpler entity-deduction settings, introducing uncertainty successfully mitigates the premature elimination of correct hypotheses caused by strict binary pruning.

However, on the more complex \textit{Detective Cases} benchmark, CA-BED continues to outperform the modified UoT variant (46.0\% vs. 33.0\%). This indicates that while likelihood smoothing improves robustness against early pruning errors, relying on hard binary partitions remains a limiting factor in domains where evidence is inherently ambiguous or distributed across multiple hypotheses.

\subsection{RQ3: Can answer planning improve information acquisition?}

In practical dialog settings, responses are not limited to binary \textit{Yes/No} answers. However, modeling open-ended or free-form responses can make exact Bayesian updates intractable. To address this, CA-BED is extended with an answer-planning mechanism. Rather than restricting queries to binary form or relying on post-hoc semantic matching, the agent generates multiple-choice questions with a discrete set of mutually exclusive candidate answers. This formulation enables tractable likelihood estimation over a controlled categorical space.

As shown in Table \ref{tab:main_results}, this extension yields mixed results depending on the complexity of the domain. In \textit{Detective Cases}, the Top-1 accuracy decreases from $46.0\%$ to $38.0\%$. Qualitative analysis suggests that this degradation arises from the LLM not being able to design mutually exclusive candidate answers in the open domain of a murder mystery, causing mismatched selections and, consequently, inaccurate likelihood estimates. However, in \textit{20 Questions}, the improvement is substantial, with accuracy increasing from $82.9\%$ to $89.2\%$. Additionally, the use of categorical answers allows the agent to extract more information per interaction, leading to shorter conversations on average, with mean length decreasing from $12.8$ to $8.8$ turns.

\subsection{RQ4: Does CA-BED Increase Information Gain In Unsuccessful Tasks?}

Prior work on information-seeking strategies \citep{hu2024uncertaintythoughtsuncertaintyawareplanning,chopra2025feedbackawaremontecarlotree,choudhury2025bedllmintelligentinformationgathering} commonly evaluate performance using overall success rate, which may not fully reflect the extent to which a model isolates informative evidence during interaction. To better separate these effects, we re-evaluate CA-BED, UoT, and Direct Prompting using two complementary metrics: (i) Top-3 accuracy, which measures whether the correct hypothesis remains among the most probable candidates in the agent's belief state, and (ii) Post-Dialog Success Rate (PD-SR), obtained by providing the full conversation transcript to an external reasoner (DeepSeek-V3.2 Reasoner) to produce a final prediction.

\begin{center}
\centering
\captionof{table}{Relationship between conversational informativeness and downstream task success. 
\emph{Top3} (\%) = proportion where the correct answer appears in the top-3 candidates. 
\emph{Post-dialog SR} (\%) (PD-SR) = accuracy of an external reasoner (DeepSeek-V3.2 Reasoner) given each conversation transcript.}
\label{tab:informativeness}
\begin{tabular}{lcccc}
\toprule
& \multicolumn{2}{c}{Detective Cases} & \multicolumn{2}{c}{20 Questions} \\
\cmidrule(lr){2-3} \cmidrule(lr){4-5}
Method & Top3$\uparrow$ & PD-SR$\uparrow$ & Top3$\uparrow$ & PD-SR$\uparrow$ \\
\midrule
Direct Prompting & 81.6          & \textbf{94.9} & \textbf{87.4} & \textbf{94.6} \\
UoT              & 29.0          & 36.0          & 60.4          & 62.7 \\
CA-BED           & \textbf{90.0} & 43.0          & \textbf{87.4} & 85.6 \\
\bottomrule
\end{tabular}
\end{center}

As shown in Table~\ref{tab:informativeness}, CA-BED achieves strong performance under the Top-3 metric. In \textit{Detective Cases}, despite the small hypothesis space (five suspects), CA-BED places the correct answer within the Top-3 in $90.0\%$ of runs, compared to $81.6\%$ for Direct Prompting and $29.0\%$ for UoT. In \textit{20 Questions}, CA-BED attains a Top-3 accuracy of $87.4\%$, matching Direct Prompting and exceeding UoT ($60.4\%$).

However, the post-dialog evaluation reveals a divergence between internal belief quality and downstream interpretability. When an external reasoner evaluates the transcripts, Direct Prompting achieves PD-SR values of $94.9\%$ on \textit{Detective Cases} and $94.6\%$ on \textit{20 Questions}, whereas CA-BED yields lower PD-SR, particularly on \textit{Detective Cases} ($43.0\%$). One contributing factor may be differences in query formulation. Direct Prompting often includes explicit hypothesis verification (e.g., direct guesses of specific entities), which can simplify the final inference step for an external model once confirmed. In contrast, CA-BED relies more heavily on attribute-based queries that progressively refine the hypothesis space. While this supports efficient probabilistic narrowing, the resulting transcripts may contain less explicit confirmation of the final answer, requiring the external reasoner to infer the solution from accumulated attributes.

\section{Limitations}

Our evaluation is restricted to settings with an explicit, discrete hypothesis set. This assumption is natural for entity-deduction tasks such as \textit{20 Questions} and \textit{Detective Cases}, but it limits the direct applicability of CA-BED to fully open-domain conversations where the relevant hypothesis space may be large, implicit, or only partially specified. Although the method can in principle be combined with dynamically generated candidate sets, we do not evaluate that setting here.

The experimental scope is also narrow in two additional respects. First, we study only two benchmark families, both of which are structured information-seeking games rather than real deployed applications. Second, the main experiments rely on model-based interactions, with DeepSeek-V3.2 Chat serving as the primary Questioner and Answerer. Appendix~\ref{sec:appendix_questioners} broadens the Questioner analysis, but broader cross-model, cross-domain, and human-in-the-loop validation remains future work.

CA-BED also inherits limitations from its LLM-based likelihood estimation. The quality of the belief updates depends on whether the model can assign well-calibrated probabilities and, in the answer-planning setting, generate candidate answers that are genuinely distinct and mutually exclusive. The degradation observed for answer planning on \textit{Detective Cases} illustrates that failures in answer-space design or likelihood estimation can directly reduce downstream accuracy.

Finally, the method incurs substantial inference-time overhead relative to direct prompting. Even with aggressive caching, CA-BED requires repeated candidate generation, likelihood estimation, and lookahead tree expansion, which increases both latency and cost. In practice, this trade-off may matter in interactive settings where response time or budget is constrained, and further work is needed to improve computational efficiency without sacrificing performance.

\section{Conclusion}

We presented Conversation-Aware Bayesian Experimental Design (CA-BED), a
probabilistic framework for inference-time dialog planning. CA-BED unites
Bayesian Experimental Design with LLM-based likelihood estimation to optimize
question selection across multiple conversational turns. By modeling uncertainty
and anticipating possible answers, it achieves more robust and efficient
information gathering. On entity-deduction benchmarks, CA-BED improved success
rates by an average of 21.8\% absolute over direct prompting and surpassed planning
methods such as UoT, requiring an average increase of only 1.8 conversational turns.
These results demonstrate that probabilistic reasoning over uncertainty offers a
principled path toward more reliable interactive reasoning in LLMs.

\bibliography{iclr2026_conference}
\bibliographystyle{iclr2026_conference}

\appendix
\section{Questioner Model Ablation}
\label{sec:appendix_questioners}

To assess whether the gains reported in the main paper depend on the specific Questioner model, we repeated the \textit{20 Questions} evaluation with \textbf{GPT-5.4-nano} and \textbf{Gemini-3.1-flash-lite} as alternative Questioners. We report these runs in the appendix, rather than the main tables, because these models were substantially more expensive to evaluate at the scale of our full benchmark suite, whereas DeepSeek-V3.2 Chat was more cost-efficient for complete experiments. In all cases, we reuse the same method hyperparameters described in Section~4.3. Table~\ref{tab:questioner_models} reports Top-1 accuracy, Top-3 accuracy, mean conversation length (MC), and mean successful conversation length (MSC) for these additional runs.

\begin{center}
\centering
\captionof{table}{Additional \textit{20 Questions} results under alternative Questioner models. Top-1 and Top-3 report Wilson's 95\% Confidence Intervals [in brackets]. MC and MSC report mean $\pm$ SEM.}
\label{tab:questioner_models}
\resizebox{\textwidth}{!}{%
\begin{tabular}{llcccc}
\toprule
Questioner & Method & Top-1$\uparrow$ & Top-3$\uparrow$ & MC$\downarrow$ & MSC$\downarrow$ \\
\midrule
\multirow{3}{*}{GPT-5.4-nano} & Direct Prompting & 28.8 [21.2, 37.9]                   & 56.8 [47.5, 65.6]                   & \textbf{11.1 $\pm$ 0.46} & \textbf{6.8 $\pm$ 0.33} \\
                                 & UoT (0.7, 0.8)   & 78.4 [69.8, 85.0]                   & 83.8 [75.8, 89.5]                   & 13.5 $\pm$ 0.40          & 12.6 $\pm$ 0.41 \\
                                 & CA-BED           & \textbf{82.0} [\textbf{73.8, 88.0}] & \textbf{86.5} [\textbf{78.9, 91.6}] & 13.9 $\pm$ 0.41          & 12.6 $\pm$ 0.37 \\
\midrule
\multirow{3}{*}{Gemini-3.1-flash-lite} & Direct Prompting & 61.3 [52.0, 69.8]                   & 86.5 [78.9, 91.6]                   & \textbf{10.5 $\pm$ 0.40} & \textbf{8.5 $\pm$ 0.28} \\
                                        & UoT (0.7, 0.8)   & 75.7 [66.9, 82.7]                   & 81.1 [72.8, 87.3]                   & 15.0 $\pm$ 0.39          & 13.8 $\pm$ 0.41 \\
                                        & CA-BED           & \textbf{82.9} [\textbf{74.8, 88.8}] & \textbf{90.1} [\textbf{83.1, 94.4}] & 13.3 $\pm$ 0.40          & 12.1 $\pm$ 0.35 \\
\bottomrule
\end{tabular}%
}
\end{center}

The additional Questioner analysis shows that CA-BED remains robust when the model generating questions is changed. With GPT-5.4-nano, direct prompting is weak ($28.8\%$ Top-1), whereas both planning-based methods recover strong performance and CA-BED achieves the best accuracy at $82.0\%$. This suggests that explicit information-seeking structure is especially helpful for smaller or less capable Questioners.

With Gemini-3.1-flash-lite, the direct prompting baseline is much stronger ($61.3\%$ Top-1), yet CA-BED still achieves the best Top-1 and Top-3 performance at $82.9\%$ and $90.1\%$, respectively. UoT (0.7, 0.8) remains competitive under both alternative Questioners but consistently trails CA-BED, indicating that soft probabilistic updates combined with Bayesian question selection continue to provide a measurable benefit. As in the main experiments, these gains come at the cost of longer dialogs than direct prompting.

\end{document}

%% file: math_commands.tex
\usepackage{amsmath,amsfonts,bm}

\def\eqref#1{equation~\ref{#1}}

\def\1{\bm{1}}

\DeclareMathAlphabet{\mathsfit}{\encodingdefault}{\sfdefault}{m}{sl}
\SetMathAlphabet{\mathsfit}{bold}{\encodingdefault}{\sfdefault}{bx}{n}

